\documentclass[11pt,a4paper]{article}
\usepackage[nohyperref]{emnlp2018}
\usepackage{times}
\usepackage{latexsym}
\usepackage{url}
\usepackage{times}
\usepackage{latexsym}
\usepackage{pgf}
\usepackage{tikz}
\usepackage{amsmath}
\usepackage{booktabs}
\usepackage{CJKutf8}
\usepackage{graphicx}
\aclfinalcopy
\newcommand{\themainpapercontent}[1]{#1}
\newcommand{\thesupplementarycontent}[1]{}

\newcommand{\out}[1]{}

\colorlet{darkred}{red!50!black}
\usetikzlibrary{decorations,arrows,shapes}

\title{Robust Chinese Word Segmentation with Contextualized Word Representations}

\author{Yung-Sung Chuang\\
 Department of Electrical Engineering, National Taiwan University \\
 {\tt b05901033@ntu.edu.tw} }

\date{}

\begin{document}
\maketitle

\themainpapercontent{

\begin{abstract}
  In recent years, after the neural-network-based method was proposed, the accuracy of the Chinese word segmentation task has made great progress. However, when dealing with out-of-vocabulary words, there is still a large error rate.

We used a simple bidirectional LSTM architecture and a large-scale pretrained language model to generate high-quality contextualize character representations, which successfully reduced the weakness of the ambiguous meanings of each Chinese character that widely appears in Chinese characters, and hence effectively reduced OOV error rate. State-of-the-art performance is achieved on many datasets.
\end{abstract}

\begin{CJK}{UTF8}{bsmi}
\section{Introduction}

For Chinese word segmentation task, many works have explored novel neural-network-based methods to improve the accuracy of segmentation prediction \cite{P14-1028,ma-hinrichs:2015:ACL-IJCNLP,DBLP:conf/acl/ZhangZF16, DBLP:journals/corr/LiuCGQL16, cai2017fast, DBLP:journals/corr/abs-1711-04411, ma2018}.

However, most of them used the context-independent character representations, or used information about neighboring words, such as using bigram embedding as input \cite{ma2018}. These context-independent character representations may cause weaknesses in the entire word segmentation model. Because in Chinese, a large number of characters have different meanings in different words, hence their true meaning cannot be determined unless they are combined together with other characters.



When the context-independent word representations are applied to the word segmentation model, the model may be able to memorize all the combinations of characters that should be combined into words if all the words have been identified in the training set, even if these characters are not truly represented by the vectors. However, if it is an out-of-vocabulary word that only appears in the testing set, this becomes a challenge for the model.

In the discussion of previous works \cite{ma2018,HUANGChang-ning:8}, the out-of-vocabulary issue was blamed for being one of the main sources of error rates. This issue will have a great impact on the practicality of using the trained model in practical applications. Because the proportion of OOV issue is relatively small in the dataset, a simple neural-network architecture may achieve a good accuracy rate in these datasets, but it does not perform well in practical applications.

However, after the emergence of contextualized word representations \cite{Peters2018}, this weakness may be effectively reduced. In the following paragraphs, we will compare the significant advances in contextualized word representations to reducing the OOV error rate.

\section{Model}

\begin{figure}
\centering
\includegraphics[height=4cm]{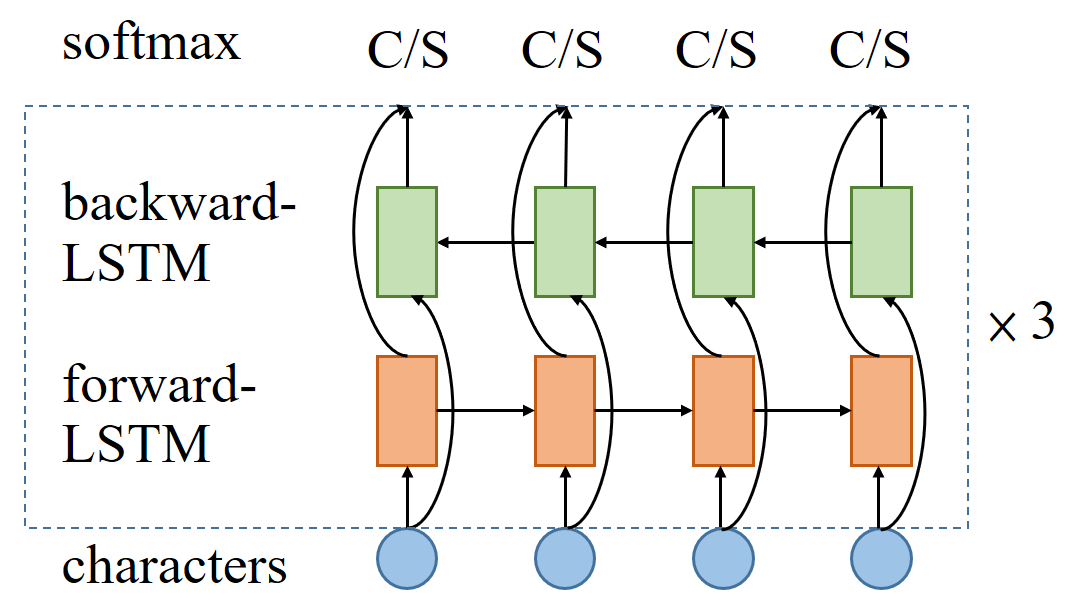}
\label{fig_models}
\caption{
Bi-LSTM models: Blue circles are input (from ELMo) embeddings.  Orange and green squares are LSTM cells, repeated for three times. C/S is a 2-way softmax to determine \texttt{continue} or \texttt{separate} at every time step.
}
\end{figure}

\subsection{Bi-LSTM model}

Our model is based on long short-term memory neural networks. Different from previous state-of-the-art model \cite{ma2018} with stacked LSTMs, we use non-stacked, three layers of Bi-LSTMs(see Figure 1). The input vectors are from ELMo embedding, which will be introduced in the following section. The output softmax layer predicts two possible state: \texttt{continue} or \texttt{separate} from transition based model. The meaning of \texttt{continue} state is to wait for next character to append, while the meaning of \texttt{separate} state is to combine the currently collected characters into a word. This operation is similar to the method used in  \cite{DBLP:conf/acl/ZhangZF16}. An example is demonstrated in Table 1.

\begin{table}
\centering
\begin{tabular}{cccccc}
\toprule
0　1 & 1 & 1 & 0　1 & 0　1 & 0　0　1 \\
\underline{歡 迎} & \underline{來} & \underline{到} & \underline{台 灣} & \underline{大 學} & \underline{電 機 系}  \\
\bottomrule
\end{tabular}
\caption{
An easy example for the output of transition based model. "1" represents "separate", and "0" represents "continue".
}
\end{table}

There are some remained hyperparameters: 300 units for LSTM hidden size, 1024 for the dimension of ELMo embedding, 0.33 for the dropout rate at the output of each LSTM layer except the top layer. We applied cross entropy to compute the loss of the softmax output with true label. We use Adam \cite{adam2014} to optimize the model with learning rate $\alpha = 10^{-3}$, $\beta_1 = 0.9$ ,$\beta_2 = 0.999$, $\epsilon = 10^{-8}$.

\subsection{ELMo}

Different from traditional word embedding concepts, in ELMo \cite{Peters2018}, word representation is a function of the sentence. That is, the vector representation of every word is not determined until the whole sentence is constructed, and thus each vector representation is contextualized. To get the contextualized representation, they use a bidirectional language models (biLM) and have it trained on large corpus.. The forward LM models the probability of token $t_k$ given the history $(t_1, ..., t_{k-1})$:

$$p(t_1, t_2, \ldots, t_N) = \prod_{k=1}^N p({t_k} \mid t_1, t_2, \ldots, t_{k-1}).$$

A backward LM runs in reverse, predicting the previous token given the future context:

$$p(t_1, t_2, \ldots, t_N) = \prod_{k=1}^N p(t_k \mid t_{k+1}, t_{k+2}, \ldots, t_N).$$

Starting from computing a context-independent token representation $x^{LM}_
k$ (via a CNN over characters) for each token $t_k$, we pass $x^{LM}_
k$ through two layers of Bi-LSTMs. At each position k, we get a context-dependent representation $\overrightarrow{\mathbf{h}}^{LM}_{k,1}$ and $\overrightarrow{\mathbf{h}}^{LM}_{k,2}$ from the two layers of forward LSTMs, as well as $\overleftarrow{\mathbf{h}}^{LM}_{k,1}$ and $\overleftarrow{\mathbf{h}}^{LM}_{k,2}$ from the two layers of backward LSTMs. The outputs of the top layers ($\overrightarrow{\mathbf{h}}^{LM}_{k,2}$, $\overleftarrow{\mathbf{h}}^{LM}_{k,2}$) are used to predict next/previous token respectively by performing a Softmax layer with tied parameters for both directions. Finally, we jointly maximize the log likelihood of both the forward and backward directions:

\[
\begin{split}
\sum_{k=1}^N \left( \right. & \log p({t_k} \mid t_1, \ldots, t_{k-1}; \Theta_x, \overrightarrow{\Theta}_{LSTM}, \Theta_s) \\
+  & \log p({t_k} \mid t_{k+1}, \ldots, t_{N}; \Theta_x, \overleftarrow{\Theta}_{LSTM}, \Theta_s)
\left. \right).
\end{split}
\]

To get the final word representation from the pretrained biLM for the downstream task, we compute:
\\
\\
$\mathbf{ELMo}_k = \gamma (s_0 \mathbf{x}^{LM}_{k} + s_1 \mathbf{h}^{LM}_{k,1} + s_2 \mathbf{h}^{LM}_{k,2})$
\\
\\
where $\gamma$ and $s_i$ can be adjusted when training on different downstream tasks. $\mathbf{s}_{i}$ are softmax-normalized weights and the scalar parameter $\gamma$ allows the task model to scale the entire ELMo vector. 

In our task, we train the ELMo model with the sentences separated into characters, in order to get better character representations. This operation will make the character CNNs in front of the ELMo model degenerate into some simple DNNs. However, by doing this, we can maintain the consistency between training and testing the ELMo model, and get more appropriate representations for every character.


\begin{table*}
  \centering
  \scalebox{0.92}{
  \begin{tabular}{l|ccccccc}
   &{} AS & CITYU &      MSR  &    PKU  & UD \\
   \hline
    CKIP & 97.7 & 94.3 & 92.0 & 93.9 & 91.2 \\
    Jieba & 87.1 & 86.8 & 86.5 & 87.6 & 87.6 \\
   \hline
    \newcite{DBLP:journals/corr/LiuCGQL16}   & --- & --- &  97.3 & 96.8  \\                 
    \newcite{yang2017neural}     & 95.7    & 96.9 &  97.5 & 96.3   & --- \\
    \newcite{zhou2017word}       & ---     & ---  &  97.8 & 96.0   & --- \\
    \newcite{cai2017fast}        & ---     & 95.6 &  97.1 & 95.8   & --- \\
    \newcite{chen2017adversarial}& 94.6    & 95.6 &  96.0 & 94.3   & --- \\
    \newcite{K17-3015}           & ---     & ---  &  ---  & ---    & 94.6 \\
    \newcite{DBLP:journals/corr/abs-1711-04411}   & ---     & ---  & 98.0  & 96.5  & --- \\
     \newcite{ma2018}        & 96.2 & 97.2 & 98.1  & 96.1    & 96.9 \\
    \hline
    Ours (baseline model) & 97.5 & 97.1 & 98.4 & 96.5 & 95.7 \\
    Ours (generally pretrained ELMo) & 97.5 & 97.8 & 98.1 & 96.8 & 97.0 \\
    Ours (character level ELMo) & \bf98.0 & \bf98.6 & \bf98.7 & \bf97.7 & \bf98.3 \\

  \end{tabular}
  }
  \caption{
    \label{tab_state_of_the_art}
    The state of the art performance on different datasets. We also compare with existing word segmentation toolkits Jieba and CKIP \cite{ckip}. 
  }
\end{table*}

\begin{table}
  \centering
  \scalebox{0.92}{
  \begin{tabular}{l|rrr}
   &{} Train & Test \\
   \hline
    AS    & 5,449,581 &  122,610 \\                 
    CTIYU & 1,455,630 &  40,936 \\
    MSR   & 2,368,391 &  106,873 \\
    PKU   & 1,109,947 &  104,372 \\
    UD    & 111,271  &  12,012 \\
   \hline
      \end{tabular}
  }
  \caption{
    \label{statistics}
    Statistics of the dataset.
  }
\end{table}

\begin{table*}
\centering
\begin{tabular}{l|rrrrr}
  &  AS & CITYU & MSR & PKU & UD \\ \hline
OOV \% & 4.2 & 7.3 & 2.6 & 3.5 & 11.9 \\ \hline
Accuracy \% (baseline model) & 90.3 & 92.7 & 93.4 & 89.2 & 91.0 \\
Accuracy \% (char-level ELMo) & \bf91.3 & \bf95.6 & \bf93.7 & \bf93.2 & \bf95.5 \\ 
\end{tabular}
\caption{
\label{test_set_oov} Testing set OOV rate, together with the accuracy achieved on OOV set by our baseline model and our main model, respectively.
}
\end{table*}

\begin{table*}
\centering
\begin{tabular}{l|rrrrr}
  &  AS & CITYU & MSR & PKU & UD \\ \hline
OOV \% & 4.2 & 7.3 & 2.6 & 3.5 & 11.9 \\ \hline
ratio of words with digits in OOV \% & 7.7 & 2.6 & \bf20.6 & 8.7 & 6.1 \\
ratio of words length $\geq$ 5 in OOV \% & 10.3 & 5.2 & \bf35.6 & 6.6 & 5.9 \\ 
\end{tabular}
\caption{
\label{oov_stc} Testing set OOV rate, together with ratio of OOV with digits and ratio of OOV with long words, respectively.
}
\end{table*}

\section{Experiments}

\subsection{Data}
We run our experiments on datasets of SIGHAN 2005 bake-off task (Emerson, 2005) and a dataset of Chinese Universal Treebank (UD) from the Conll2017 shared task \cite{zeman-EtAl:2017:K17-3}, both with the official data split.
Table~\ref{statistics} shows statistics of each data set.
We randomly select $5\%$ training set as the validation set while training. We convert MSR and PKU dataset to traditional Chinese.
Following \newcite{ma2018}, we train and evaluate a model for each of the dataset, rather than train one model on the union of all dataset. We also convert all digits, punctuation and Latin letters to half-width for maintaining the consistency between the training and testing sets.

For preparing the corpus for training embeddings, we made a union corpus composed of five training sets (about 22.8M tokens) and randomly sampled sentences from Chinese gigaword (about 7.8M tokens), exactly two million sentences in total. All the sentences are separated into characters. We use this corpus to train a simple skip-gram Word2Vec embedding \cite{mikolov2013efficient} for our baseline model. And we train an ELMo model for 10 epochs on the same corpus, used by our main model. 

\subsection{Results}
We compare our model with the state-of-the-art results from recent neural network based models, together with existing word segmentation toolkits Jieba and CKIP \cite{ckip} (see Table~\ref{tab_state_of_the_art}). We achieve the best results on all datasets. 

We can notice that our baseline model with pretrained skip-gram embeddings performs good enough to be competitive with the previous state-of-the-art results. However, when discussing the OOV error rate in the next section, we can see the significant difference between the baseline model and the main model. We also tested a generally pretrained Traditional Chinese ELMo model provided by HIT-SCIR \cite{che-EtAl:2018:K18-2}, trained on Wikipedia, which performed slightly worse than our main model.

\subsection{Progress on OOVs}

In all the datasets, there is a certain proportion of words only shown in the testing set, but not shown in the training set, namely, out-of-vocabulary (OOV). Thus, the model can not just remember the combinations of all the words in the training set to perform well on OOV words. It must have the ability to generalize the knowledge to combine characters into words on the testing set.

In Table~\ref{test_set_oov} we show the statistics of the OOV rate among these datasets and compare the performance between our main model and baseline model. We can see that even though the baseline model can have a good accuracy nearly 97\% on the whole testing sets of these datasets, the accuracy computed only on OOV words is still stuck in nearly 91\%. However, our main model still maintains relatively good performance on OOV words outperforming the baseline model with at most about 5\% accuracy, showing the robustness of our model.

Nevertheless, we notice that the accuracy of MSR dataset is not progressed much compared to the baseline model. We observe that the MSR dataset has more consistency between the training set and testing set with low OOV rate (2.6\%), as mentioned by \cite{ma2018}. However, many OOVs come from strings with digits in Chinese, such as "15類", "458家", "2686個", "去年6月", "4月19日". Other OOVs largely come from name entities, such as "蚌埠檸檬酸廠", "靖邊天然氣淨化廠", "長江水文局", "中共湖北省委", all of them are tied to be single words. In these situations, our main model can not make full use of its advantages, because these words are more likely to be segmented in that way by human defined rules instead of its natural semantic meaning. When dealing words with digits, it is easy for both model; when dealing longer name entities, our model may segment them into smaller pieces while our baseline model also does that. Hence, testing on MSR dataset does not cause obviously different results comparing to the baseline model. 

To prove this assumption is true, we made some statistics on the testing data (see Table~\ref{oov_stc}). We counted all words with digits in it in the OOV set of each testing data. The MSR dataset has about 20\% OOV words with digits in it, while other datasets only have the ratio of less than 10\% OOV words with digits in it. We can not identify all name entities in OOV set directly, however, we noticed that long words are more likely to be name entities in Chinese, especially when words are not shorter than 5 characters. As a result, we counted all words with their length $\geq$ 5 (see Table~\ref{oov_stc}), and prove that MSR indeed have larger (about 36\%) OOV words with their length $\geq$ 5, while other datasets only have the ratio of less than 10\% OOV words with their length $\geq$ 5.


\subsection{Other discussions}

When looking into some error segmented words, we find lots of words are ambiguous to decide whether to segment it into small segments, which is the thing our model usually do. For example, "老歌" (old song), "多次" (many times), "抗癌" (against cancer) can be segmented to smaller units as "老 歌", "多 次", "抗 癌" without breaking their semantic meaning. However, our evaluation metrics are not taking this into account. We only care how data is annotated, so that all this kind of errors contribute to the error rate, which is not a fair way of evaluating the performance. On the other side, human evaluation can be more accurate, but it is more expensive and time-consuming. The further improvement on this issue may be useful for Chinese word segmentation task.

\end{CJK}

\section{Conclusion}

In this work, we used a simple Bi-LSTM model together with embedding from language models (ELMo), achieving state-of-the-art performance on many Chinese word segmentation  datasets. Also, we found that our model is more robust when dealing with the out-of-vocabulary issue, showing that our model can perform better, when applied into practical applications.

\newpage 
}

\bibliography{emnlp2018}
\bibliographystyle{acl_natbib_nourl}

\clearpage

\thesupplementarycontent{

\appendix

\section{Supplemental Material}
\label{sec:supplemental}

}

\end{document}